\documentclass[conference]{IEEEtran}
\IEEEoverridecommandlockouts
\usepackage{cite}
\usepackage{amsmath,amssymb,amsfonts}
\usepackage{algorithmic}
\usepackage{algorithm}
\usepackage{graphicx}
\usepackage{textcomp}
\usepackage{xcolor}
\usepackage{subfigure}
\usepackage{multirow}
\usepackage{url}
\def\BibTeX{{\rm B\kern-.05em{\sc i\kern-.025em b}\kern-.08em
    T\kern-.1667em\lower.7ex\hbox{E}\kern-.125emX}}
\begin{document}
\title{Asynchronous Federated Learning with Incentive Mechanism Based on Contract Theory\\}

\author{\IEEEauthorblockN{
Danni Yang\textsuperscript{1},\
Yun Ji\textsuperscript{1},
Zhoubin Kou\textsuperscript{1},\
Xiaoxiong Zhong\textsuperscript{1,2}, and 
Sheng Zhang\textsuperscript{1,*}}
\IEEEauthorblockA{\textsuperscript{1}Shenzhen International Graduate School, Tsinghua University, Shenzhen, 518055, China}
\IEEEauthorblockA{\textsuperscript{2}Peng Cheng Laboratory, Shenzhen 518000, P.R. China}
\IEEEauthorblockA{\textsuperscript{*}Corresponding author: Sheng Zhang, email: zhangsh@sz.tsinghua.edu.cn}}

\maketitle

\begin{abstract}
To address the challenges posed by the heterogeneity inherent in federated learning (FL) and to attract high-quality clients, various incentive mechanisms have been employed. However, existing incentive mechanisms are typically utilized in conventional synchronous aggregation, resulting in significant straggler issues. In this study, we propose a novel asynchronous FL framework that integrates an incentive mechanism based on contract theory. Within the incentive mechanism, we strive to maximize the utility of the task publisher by adaptively adjusting clients' local model training epochs, taking into account factors such as time delay and test accuracy. In the asynchronous scheme, considering client quality, we devise aggregation weights and an access control algorithm to facilitate asynchronous aggregation. Through experiments conducted on the MNIST dataset, the simulation results demonstrate that the test accuracy achieved by our framework is 3.12\% and 5.84\% higher than that achieved by FedAvg and FedProx without any attacks, respectively. The framework exhibits a 1.35\% accuracy improvement over the ideal Local SGD under attacks. Furthermore, aiming for the same target accuracy, our framework demands notably less computation time than both FedAvg and FedProx.
\end{abstract}


\section{Introduction}
With the advancement of computing technology and the widespread integration of artificial intelligence (AI), a plethora of machine learning-based services have emerged. To address concerns related to privacy and efficiency in centralized data processing, a novel distributed machine learning paradigm known as federated learning (FL) has gained prominence, which significantly influences the development and application of machine learning at edge devices. However, within the wireless network scenario, FL encounters two primary challenges. 1) Willingness to Participate: engaging in FL consumes device's computing and communication resources, in addition to security and privacy risks. Clients require adequate incentives to offset the costs associated with their involvement. Task publishers, likewise, must enhance rewards to attract high-quality clients. 2) Heterogeneity: FL suffers from system and data heterogeneity, resulting in non-identically and independently distributed (non-IID) data distribution and varying resource allocation. Consequently, this leads to distinct computing delays and learning quality among devices.

To tackle these two primary challenges in FL, the incentive mechanism can be introduced \cite{4,5}. For the Parameter Server (PS), accurate assessment of each client's contribution is essential to maximize benefits while minimizing rewards. On the clients' side, achieving a balance between benefits and losses hinges on the effectiveness of the incentive mechanism. However, ensuring its effectiveness is challenged by information asymmetry issues stemming from stringent privacy concerns. Contract theory emerges as a promising solution due to its self-disclosure mechanism, which can enable clients to receive rewards commensurate with their contributions. Nevertheless, existing incentive mechanisms in FL are primarily tailored for synchronous aggregation, a process carried out after all clients have completed their local training steps. However, due to FL's inherent heterogeneity, synchronous aggregation exacerbates straggler issues and poses the risk of losing the features of crucial data instances due to bottlenecks. Recent works proposed asynchronous schemes to address straggler issues, such as full asynchronous \cite{14} and time-triggered aggregation asynchronous \cite{2}. Regrettably, they overlook the willingness to participate among high-quality clients.

Inspired by this, we propose the integration of an incentive mechanism into the asynchronous federated learning (FL) system, introducing two improvements to this fusion. Firstly, conventional incentive mechanisms are predominantly designed considering the utility derived from clients' local data quality \cite{5} (i.e., data quantity and distribution). However, they often overlook the utility stemming from clients' local model training epochs, which significantly impacts the accuracy of the global model. Traditional incentive mechanisms mandate all clients to execute the same local training epochs, which is unrealistic under the resource constraint of system heterogeneity. Hence, it is rational to permit variable local training epochs to be performed locally among clients. Secondly, within the asynchronous setting, a client with higher data quality and shorter training time should be rewarded more. This approach incentivizes clients to accelerate their training process. To impartially evaluate contributions from diverse clients and motivate high-quality clients to expedite their training tasks, we integrate an incentive mechanism into the asynchronous FL process. The main contributions are summarized as follows:
\begin{itemize}
\item Based on contract theory, we propose an asynchronous federated learning framework combined incentive mechanism. In this paper, we divide the clients into $N$ levels based on their data quality. For the task publisher, to make contracts for different levels of clients, we formulate the optimization problem of maximizing the utility function which is related to the FL latency and accuracy. By solving it we obtain the number of local training epochs and rewards for each level of clients which will be incorporated into contracts.
\item After the clients finish their training based on the content of their corresponding contracts, we propose an access control algorithm for the PS to govern access decisions and determine the aggregating weights of each client. By considering the client’s level, staleness, and learning quality, this algorithm can effectively guard against access of attackers and optimize the aggregation weights to enhance accuracy.
\item We perform experiments on the MNIST dataset and compared with other baselines the proposed scheme can stimulate more high-quality model updates in asynchronous FL, and it also preserves the benefits of superior global test accuracy and training time.
\end{itemize}

\section{System Model}

\subsection{FL Problems}

We consider a wireless FL system consisting of a task publisher, a PS, and a set of $\mathcal{K}=\{1, \ldots, K\}$ clients. Device $k \in \mathcal{K}$ obtains a shared global model from the PS and starts training on its local dataset $\mathcal{D} _k=\left\{ \left( \boldsymbol{x}_{k,1},y_{k,1} \right) ,\dots ,\left( \boldsymbol{x}_{k,D_k},y_{k,D_k} \right) \right\} $ , which contains $\left|\mathcal{D}_k\right|=d_k$ data samples. $\left( {x}_{k,i},y_{k,i} \right)$ is the $i$-th input-output pair of client $k$, where $\boldsymbol{x}_{k,i}$ denotes the feature vector and $y_{k,i}$ denotes the corresponding label. The sample set of total data from $K$ devices can be expressed as $\mathcal{D}=\{\mathcal{D}_{1},\ldots,\mathcal{D}_{K}\}$ with total size  ${D}=\sum_{k=1}^{K}{d}_{k}$. FL optimizes the global loss function by computing the weighted average of local losses for each device according to
\begin{equation}\label{Eq:sys_FL_task}
    \min_{\boldsymbol{w}}\ F(\boldsymbol{w})=\sum\nolimits_{k=1}^K \frac{d_k}{D}F_k(\boldsymbol{w}),
\end{equation}
where $F_k$ is the local loss function defined as
\begin{equation}\label{Eq:sys_local_loss}
    F_k(\boldsymbol{w})=\frac{1}{d_k}\sum\nolimits_{i\in \mathcal{D}_{K}}f_i(\boldsymbol{w}),
\end{equation}
where $f_i(\boldsymbol{w})$ is the loss function of sample data $i$.

\subsection{Asynchronous FL with Periodic Aggregation}
Inspired by \cite{2}, we consider an asynchronous FL with periodic aggregation. Assuming needs $T$ communication rounds in total, the process of a round is as follows:

\textbf{(a) local model computation:} At the beginning of round $t$, the client that is going to participate in the FL task receives the global model $\boldsymbol{w}_g^t$ from the PS, and the client that did not participate in the global aggregation of the previous round cannot receive the initial global model for this round. If the client completes the computation of the round $r$ model in round $t$, the update rule of client $k$ can be formulated as
\begin{equation}\label{Eq:update rule}
    \boldsymbol{w}_k^t=\boldsymbol{w}_g^r-\eta\sum_{\tau=1}^{\tau_k}\nabla F_k(\boldsymbol{w}_k^{\tau-1}),
\end{equation}
where $\eta$ is the learning rate, $r=\max_{t^{\prime}<t}\{t^{\prime}|k\in\mathcal{K}(t^{\prime})\}+1$ indicates the last time client $k$ received the global model (i.e., the time stamp of local model), then the staleness is $t-r$. $\tau_k$ is the local training epochs of the client $k$. 

In terms of the computation delay and energy overheads incurred in this process, we let $f_k$ denote the computation resources (i.e., CPU cycle frequency) contributed to the local model training. The number of CPU cycles required to perform a data sample is $c_k$. Therefore, the computation delay for client $k$ to perform a local training epoch can be formulated as
\begin{equation}\label{Eq:Tcmp}
T_k^{cmp}={c}_{k}{d}_{k}/{f}_{k},
\end{equation}
and CPU energy consumption can be formulated as
\begin{equation}\label{Eq:Ecmp}
E_{k}^{cmp}=\xi{c}_{k}{d}_{k}{f_{k}}^{2},
\end{equation}
where $\xi$ represents the effective capacitance coefficient \cite{5}.

\textbf{(b) local model communication:} At the end of round $t$, all clients ready to upload the local models form a set $\mathcal{K}(t)$, and upload their local models by the traditional uplink transmission methods with orthogonal access technologies such as FDMA\cite{5}. The transmission rate of client $k$ is denoted as $r_k={b}_{k}\ln(1+\rho_{k}{h}_{k}/{n}_{0})$, $b_k$ is the transmission bandwidth and $\rho_k$ is the transmission power of the client $k$. $h_k$ is the channel gain of the uplink transmission between client $k$ and the PS. $n_0$ represents the independent identical distribution (IID) additive Gaussian white noise (AWGN). The transmission time of a local model update can be formulated as 
\begin{equation}\label{Eq:Tcom}
T_k^{com}=\zeta_k/({b}_{k}\ln(1+\rho_{k}{h}_{k}/{n}_{0})),
\end{equation}
where $\zeta_k$ is the size of the data upload, it is a constant for all clients. The transmission energy can be formulated as 
\begin{equation}\label{Eq:Ecom}
E_{k}^{com}=\zeta_{k}\rho_{k}/({b}_{k}\ln(1+\rho_{k}{h}_{k}/{n}_{0})).
\end{equation}

\textbf{(c) local model aggregation:} At this step, the PS aggregates local model parameters from $\mathcal{K}(t)$, the weighted average aggregation strategy is according to
\begin{equation}\label{Eq:aggregation strategy}
\boldsymbol{w}_{g}^{t+1}=\boldsymbol{w}_{g}^{t}+\sum_{k\in\mathcal{K}(t)}\alpha_{k}^{t}\Delta\boldsymbol{w}_{k}^{t},
\end{equation}
where $\alpha_k^t$ indicates the weight coefficient of client $k$ in round $t$, which satisfies $\sum_{k\in\mathcal{K}(t)}\alpha_{k}^{t}=1$. Therefore, the total time of a communication round for the client $k$ is denoted as $T_k=\tau_{k}{T}_{k}^{cmp}+{T}_{k}^{com}$, and the total energy consumption is denoted as $E_k=\tau_{k}{E}_{k}^{cmp}+{E}_{k}^{com}$.

\section{Incentive Mechanism and Access Control Algorithm}
\subsection{Overview}
\begin{figure}[t]
\centerline{\includegraphics[width=0.5\textwidth]{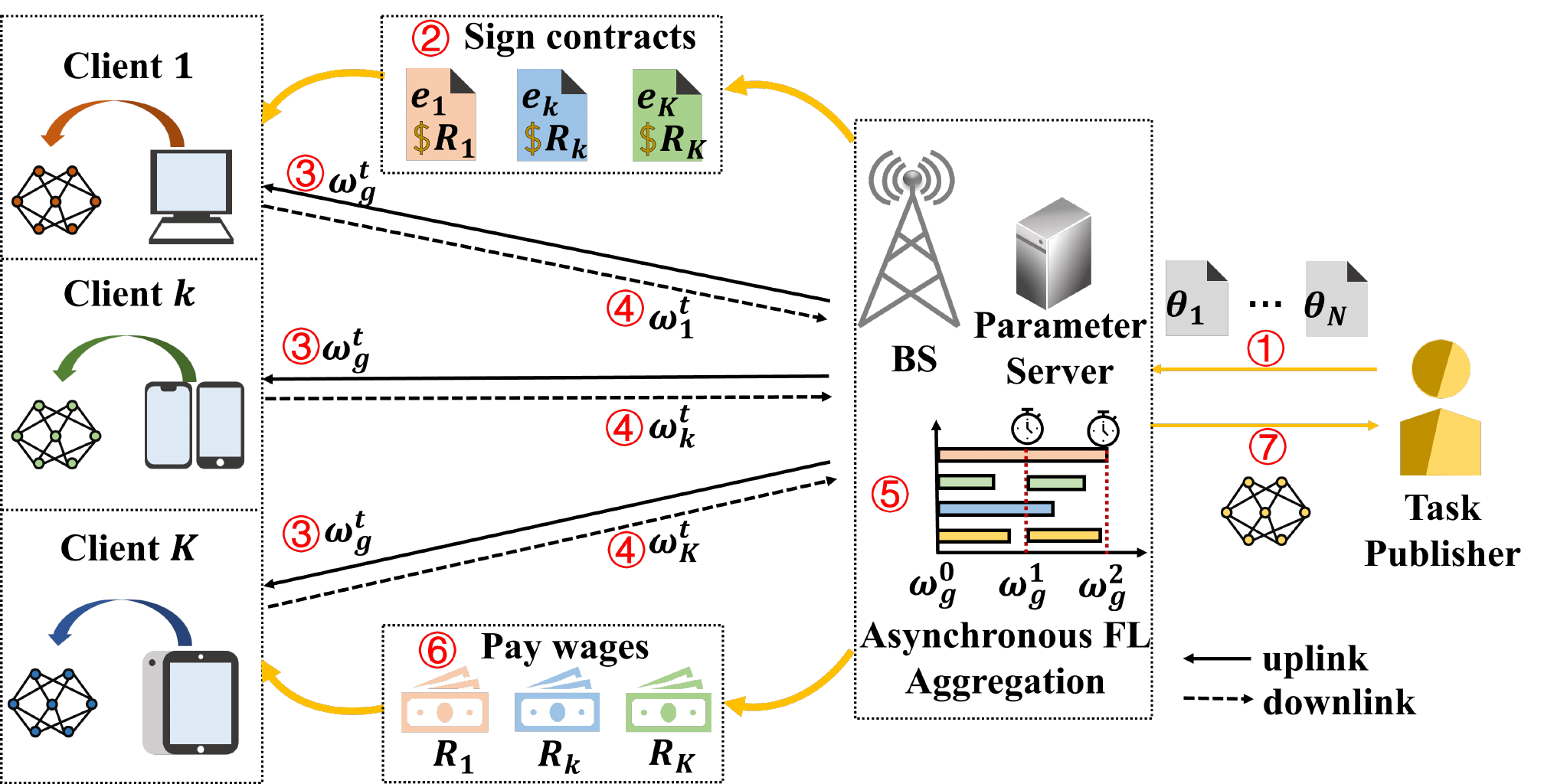}}
\caption{Asynchronous FL system with incentive mechanism.}
\label{fig:sys_fig}
\end{figure}

Our asynchronous FL architecture based on contract theory is shown in Fig.~\ref{fig:sys_fig}. It consists of seven main processes. 1) We assume that the clients can be divided into $N$ levels based on their data quality. Firstly the task publisher should publish the task and make the contracts for these $N$ levels. Each contract is obtained by maximizing the utility of PS, which is elaborately depicted in parts B and C of this section. After solving the optimization problem, the PS will get the rewards and efforts of clients on this level which is the main content of each contract, where efforts can determine the local training epochs. 2) The PS publishes all contracts to clients, and each client can sign the corresponding contract based on their own data quality level, which means that they obtain the local training epochs. 3) After that, the PS sends the global model to clients. The clients train their own local models based on the global model. 4) A client that is ready to upload the model in this round performs local model transmission. 5) The PS calls the access control algorithm to decide the access decisions and the aggregating weights. This algorithm is elaborately depicted in part D of this section. Repeat 3-5 until the asynchronous FL process is complete. 6) The PS pays clients corresponding rewards according to the contracts while considering the principle of nonpayment and backward compatibility \cite{5}. 7) The PS returns the trained global model to the task publisher.

\subsection{Preliminaries}
\begin{figure}[t]
\centerline{\includegraphics[width=0.5\textwidth]{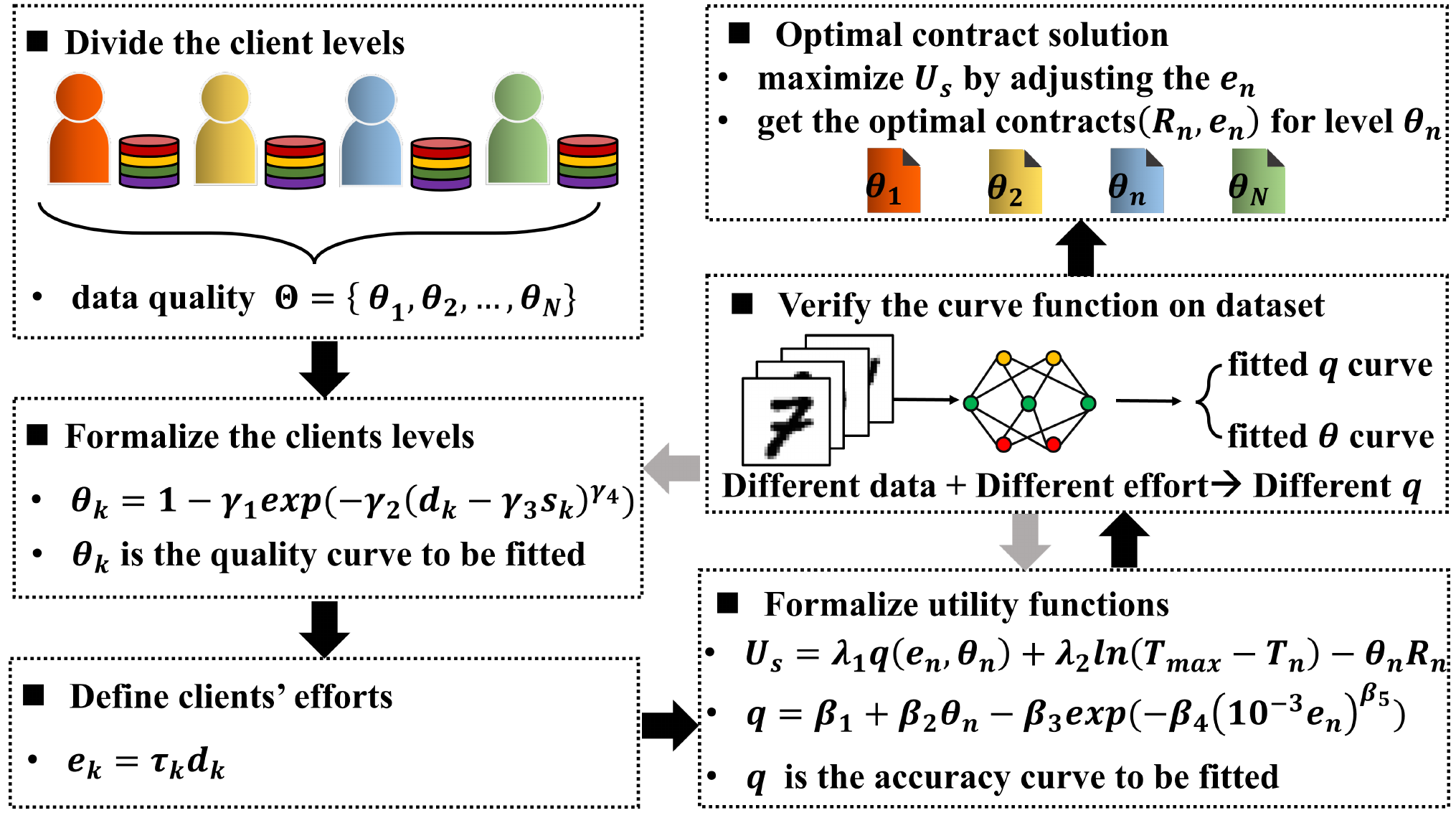}}
\caption{Contract design flow chart.}
\label{fig:liuch_fig}
\end{figure}

The contract design includes the following steps, as shown in Fig.~\ref{fig:liuch_fig}.

\textbf{a) Divide the client levels:}
The task publisher divides clients in the market into a level set $\Theta$ with $N$ levels( $\Theta=\{\theta_{1},\theta_{2},\ldots,\theta_{N}\}$ ) by their data quality, which is sorted in ascending order: $\theta_1<\cdots<\theta_n<\cdots<\theta_N,n\in\{1,\ldots,N\}$. Then, provide a set of $N$ levels of contracts $(R_n(e_n),e_n),~n\in\{1,\ldots,N\}$ that represent the relationship between clients’ rewards $R_n$ and efforts  $e_n$. Because of information asymmetry, the task publisher only knows the probability that the client belongs to a certain level through observation and statistics, which satisfies $\sum_{n=1}^{N}{p_{n}}=1$.

\textbf{b) Formalize the clients' levels:} In most works \cite{5,8}, the expression of the client level is defined by authors, which lacks scientificity and accuracy, we adopt the fitting method to determine the expression according to the data quality(i.e., data quantity and the data distribution). The data distribution is the earth mover’s distance (EMD) from the overall actual distribution. 

\emph{\textbf{Definition 1 (Data Quality):} determined by data quantity and the data distribution, which represents the client level}
\begin{equation}\label{Eq:theta}
{\theta_k=1-\gamma_1\exp(-\gamma_2(d_k-\gamma_3s_k)^{\gamma_4})},
\end{equation}
where the clients with $\theta_{k}\in(\theta_{n-1},\theta_{n}]$ belong to level-$n$. 

\textbf{c) Define clients’ efforts:} The greater the epochs $\tau_k$ and the larger the data quantity $d_k$, the more effort $e_k$ client invests.

\emph{\textbf{Definition 2 (Efforts):} the extent of the client taking their efforts}
\begin{equation}\label{Eq:ek}
e_k=\tau_kd_k.
\end{equation}

\textbf{d) Formalize utility functions:} We formalize the test accuracy and round delay, incorporating them into the task publisher's utility function, which can be formulated as
\begin{equation}\label{Eq:Us}
{{U_{{s}}(e_{{n}},R_{{n}})=\lambda_{{1}}q\left(e_{{n}},\theta_{{n}}\right)+\lambda_{{2}}\ln\left(T_{{\max}}-T_n\right)-\theta_{{n}}R_{{n}}}},
\end{equation}
where ${q}(e_n,\theta_n)$ represents the accuracy of the local model when the efforts of a client with level-$n$ is $e_n$. $T_{max}$ is the task publisher’s maximum tolerance time of FL. According to \cite{5}, $\ln\left(T_{{\max}}-T_n\right)$ represents the satisfaction of the task publisher regarding the round delay, $T_n={T_n^{com}+\frac{\tau_nd_nc_n}{fn}}={T_n^{com}+\frac{e_nc_n}{fn}}$. If the local model passes the access control algorithm, the client will be rewarded with $R_n(e_n)$. The probability of passing the algorithm is determined by the client's level $\theta_{{n}}$ \cite{8}. 

\textbf{e) Verify the curve function:} We adopt the curve fitting method in \cite{4} and formulate the test accuracy as
\begin{equation}\label{Eq:qet}
q(e_n,\theta_n)=\beta_1+\beta_2\theta_n-\beta_3\exp(-\beta_4{\left(10^{-3}e_n\right)}^{\beta_5}),
\end{equation}
where $\theta_{{n}}$ is also one of the unknowns fitted, according to the fitted value, \eqref{Eq:theta} can be determined.

The utility function of a level-$n$ client signed the corresponding contract $(R_n(e_n),e_n)$ can be formulated as
\begin{equation}\label{Eq:U_c}
\begin{aligned}
{U_c(e_n)}={\theta_nR_n}-{E_n},
\end{aligned}
\end{equation}
where $E_n=E_n^{com}+\xi{\tau_nd_nc_nf_n^2}=E_n^{com}+\xi{e_nc_nf_n^2}$.

With the groundwork described above, the task publisher can now design an optimal contract. See the following part for details.
\subsection{Optimal Contract Solution}
In the task publisher's utility function, both a higher level of client and more effort can increase the task publisher's profit, i.e., $\frac{\partial U_s}{\partial\theta_n}>0$ and $\frac{\partial U_s}{\partial e_n}>0$. The goal of the task publisher is to maximize its profit. Since the client can reject the contract for no cost, in this moral hazard problem, the following constraints must be met: 1) Individual Rational (IR) constraint and 2) Incentive Compatibility (IC) constraint.
\emph{\textbf{Definition 3 (Individual Rational)}: Only when the utility of the client participation is not less than 0, it will choose to participate.}
\begin{equation}\label{Eq:IR}
\begin{aligned}
{U_C(e_n)=\theta_nR_n-\xi{e}_nc_nf_n^2-E_n^{com}\geq0}.
\end{aligned}
\end{equation}

\emph{\textbf{Definition 4 (Incentive Compatibility)}: The contract design must guide the client to choose the contract level corresponding to its own level, and not choose other contracts.}
\begin{equation}\label{Eq:IC}
\begin{aligned}
{\theta_n}R_n-\xi{e}_nc_nf_n^2-E_n^{com}{\geq}{\theta_nR_m-\xi{e}_mc_mf_m^2-E_m^{com}}.
\end{aligned}
\end{equation}

In this study, we standardize the transmission bandwidth, channel gain, and transmission power for all clients. Furthermore, regarding computing resources, we maintain uniformity in CPU cycle frequency and CPU cycle numbers \cite{5}, i.e., $c_1=\cdots=c_n,\quad f_1=\cdots=f_n,\quad T_1^{com}=\cdots=T_n^{com},\quad E_1^{com}=\cdots=E_n^{com},\quad{n}=1, \cdots, N$. Therefore, the optimization problem can be formulated as

\begin{subequations}\label{opt1:max_Us}
    \begin{flalign}
\max_{(R_n,e_n)}U_s&=\sum_{n=1}^Np_n(\lambda_1(\beta_1+\beta_2\theta_n-
    \beta_3e^{(-\beta_4{(10^{-3}e_n)}^{\beta_5})})\notag \\
    &+\lambda_2\ln{(T_{\max}-{T_n^{com}-\frac{e_nc_n}{fn}})}-\theta_nR_n)\label{opt1:objective}\\
    {\text{s.t.}}\ \  &\theta_nR_n-\xi{e}_nc_nf_n^2-E_n^{com}\geq0, n=1, \cdots, N\label{opt1:c1} \\
    &{\theta_n}R_n-\xi{e}_nc_nf_n^2{\geq}{\theta_nR_m-\xi{e}_mc_mf_m^2},\notag \\
    &n, m=1, \cdots, N, n\neq m\label{opt1:c2} \\
    &T_{n}^{com}+\frac{e_{n}c_{n}}{f_{n}}\leq T_{\max}, n=1, \cdots, N\label{opt1:c3}
    \end{flalign}
\end{subequations}

We can simplify problem \eqref{opt1:max_Us} by performing the following transformation. 

\emph{\textbf{Lemma 1}: If and only if $\theta_n\geq\theta_m$, there must be $e_n\geq{e}_m$, so that $R_n\geq{R}_m$.}

\begin{IEEEproof}
Based on the IC constraint, we have
\begin{equation}\label{Eq:lemma_1_1}
    \begin{aligned}
{\theta_n}R_n-\xi{e}_nc_nf_n^2 { \geq } {\theta_nR_m-\xi{e}_mc_mf_m^2}
    \end{aligned}
\end{equation}
\begin{equation}\label{Eq:lemma_1_2}
    \begin{aligned}
&{\theta_m}R_m-\xi{e}_mc_mf_m^2 { \geq } {\theta_mR_n-\xi{e}_nc_nf_n^2}.
    \end{aligned}
\end{equation}

By combining \eqref{Eq:lemma_1_1} and \eqref{Eq:lemma_1_2}, we can get $(\frac1{\theta m}-\frac1{\theta n})(e_{n}-e_{m})\geq0$, i.e., $e_n\geq{e}_m$, if and only if $\theta_n\geq\theta_m$. At the same time, we have ${\theta_n}(R_n-R_m){ \geq } \xi({e}_n-{e}_m)c_nf_n^2$ according to \eqref{Eq:lemma_1_1}, i.e., $R_n\geq{R}_m$, if and only if $e_n\geq{e}_m$. Hence, the monotonicity must be satisfied.
\end{IEEEproof}

\emph{\textbf{Lemma 2}: The other levels’ IR constraints can be reduced if the level-$1$ client’s IR constraint is held.}
\begin{IEEEproof}
Based on IC constraints and the non-negativity of $R_n$, we have
\begin{equation}\label{Eq:lemma_2_1}
\begin{aligned}
{\theta_n}R_n-\xi{e}_nc_nf_n^2 { \geq } {\theta_nR_1-\xi{e}_1c_1f_1^2}
\end{aligned}
\end{equation}
\begin{equation}\label{Eq:lemma_2_2}
\begin{aligned}
{\theta_nR_1-\xi{e}_1c_1f_1^2} { \geq } {\theta_1R_1-\xi{e}_1c_1f_1^2}.
\end{aligned}
\end{equation}

Hence, all the IR constraints can be reduced as 
\begin{equation}\label{Eq:lemma_2_3}
\begin{aligned}
{\theta_1R_1-\xi{e}_1c_1f_1^2-E_1^{com}}{ \geq }0.
\end{aligned}
\end{equation}
\end{IEEEproof}

\emph{\textbf{Lemma 3}: Based on \textbf{lemma 1}, the IC constraints can be reduced as the local downward incentive compatibility (LDIC) constraints:}
\begin{equation}\label{Eq:LDIC}
\begin{aligned}
{\theta_n}R_n-\xi{e}_nc_nf_n^2 { \geq } {\theta_nR_{n-1}-\xi{e}_{n-1}c_{n-1}f_{n-1}^2},
\end{aligned}
\end{equation}
\emph{where $n=2,\cdots, N,$ and the local upward incentive compatibility (LUIC)constraints:}
\begin{equation}\label{Eq:LUIC}
\begin{aligned}
{\theta_n}R_n-\xi{e}_nc_nf_n^2 { \geq } {\theta_nR_{n+1}-\xi{e}_{n+1}c_{n+1}f_{n+1}^2},
\end{aligned}
\end{equation}
\emph{where $n=1,\cdots, N-1.$}

\begin{IEEEproof}
    According to the LDIC, we have
    \begin{equation}\label{opt2:lemma_3_1}
    \begin{aligned}
    &\theta_{n+1}R_{n+1}-\xi{e}_{n+1}c_{n+1}f_{n+1}^2 { \geq } {\theta_{n+1}R_n-\xi{e}_nc_nf_n^2}
    \end{aligned}
    \end{equation}
    \begin{equation}\label{opt2:lemma_3_2}
    \begin{aligned}
    &{\theta_n}R_n-\xi{e}_nc_nf_n^2 { \geq } {\theta_nR_{n-1}-\xi{e}_{n-1}c_{n-1}f_{n-1}^2}.
    \end{aligned}
    \end{equation}
Because of the monotonicity and \eqref{opt2:lemma_3_2}, we have
\begin{equation}\label{opt2:lemma_3_ldic_2}
\begin{aligned}
\theta_{n+1}(R_{n}-R_{n-1}){ \geq }\theta_{n}(R_{n}-R_{n-1}){ \geq }\xi({e}_n-{e}_{n-1})c_nf_n^2
\end{aligned}
\end{equation}
Combined \eqref{opt2:lemma_3_1} and \eqref{opt2:lemma_3_ldic_2}, we have
\begin{equation}\label{opt2:lemma_3_ldic_3}
\begin{aligned}
\theta_{n+1}R_{n+1}-\xi{e}_{n+1}c_{n+1}f_{n+1}^2 { \geq } {\theta_{n+1}R_{n-1}-\xi{e}_{n-1}c_{n-1}f_{n-1}^2},
\end{aligned}
\end{equation}
it can be extended downward to level-$1$. Similarly, we can prove that the LUIC can be extended upward to level-$N$.
\end{IEEEproof}

\emph{\textbf{Lemma 4}: In the case of the monotonicity, the LDIC and the LUIC can be simplified as}
\begin{equation}\label{opt2:lemma_4}
\begin{aligned}
{\theta_n}R_n-\xi{e}_nc_nf_n^2 \geq {\theta_nR_{n-1}-\xi{e}_{n-1}c_{n-1}f_{n-1}^2}
\end{aligned}
\end{equation}

\begin{IEEEproof}
For the constraint in \eqref{Eq:lemma_2_3}, the task publisher will lower the $R_1$ as possible as it can to optimize its utility function, until ${\theta_1R_1-\xi{e}_1c_1f_1^2-E_1^{com}} = 0$. Similarly, as for the LDIC, the task publisher will lower all $R$ as much as possible until ${\theta_n}R_n-\xi{e}_nc_nf_n^2 = {\theta_nR_{n-1}-\xi{e}_{n-1}c_{n-1}f_{n-1}^2}$. Because the LDIC will still hold if both $R_n$ and $R_{n-1}$ are lowered by the same amount. At the same time, ${\theta_n}R_n-\xi{e}_nc_nf_n^2={\theta_nR_{n-1}-\xi{e}_{n-1}c_{n-1}f_{n-1}^2}$ equally becomes $\theta_{n}(R_{n}-R_{n-1})=\xi({e}_n-{e}_{n-1})c_nf_n^2$. Because of the monotonicity, we have $\theta_{n}(R_{n}-R_{n-1}){ \geq }\theta_{n-1}(R_{n}-R_{n-1})$. Combined with \eqref{opt2:lemma_4}, we have
\begin{equation}\label{opt2:lemma_4_proof}
\begin{aligned}
{\theta_{n-1}}R_{n-1}-\xi{e}_{n-1}c_{n-1}f_{n-1}^2 \geq {\theta_{n-1}R_{n}-\xi{e}_{n}c_{n}f_{n}^2},
\end{aligned}
\end{equation}
which is the LUIC condition. So the LDIC and LUIC can be simplified as the above equation.
\end{IEEEproof}
\begin{algorithm}[t]
\caption{Asynchronous FL with Incentive Mechanism}
\label{alg:fedinasy}
\begin{algorithmic}[1] 
\FOR{$t=0,1, \cdots, T-1$}
    \STATE PS broadcasts $\boldsymbol{w}_\text{g}^t$ to clients $k$, $k\in\mathcal{K}(t)$;
    \FOR{$k=1, \cdots, |\mathcal{K}(t)| $ }
        \FOR{$\tau=1,\cdots, \tau_k$}
            \STATE 
            $\boldsymbol{w}_k^t\leftarrow\boldsymbol{w}_g^r-\eta\sum_{\tau=1}^{\tau_k}\nabla F_k(\boldsymbol{w}_k^{\tau-1})$;
            \IF{$train time<\Delta t$}
                \STATE $\mathcal{K}(t+1)\leftarrow\mathcal{K}(t+1)+\{k\}$;
            \ENDIF
        \ENDFOR 
    \ENDFOR
    \STATE PS calls Access Control Algorithm (\textbf{Algorithm \ref{alg:Access Control Algorithm}})
    \STATE $\boldsymbol{w}_{g}^{t+1}\leftarrow\boldsymbol{w}_{g}^{t}+\sum_{k\in\mathcal{K}(t)}\alpha_{k}^{t}\Delta\boldsymbol{w}_{k}^{t}$
\ENDFOR
\end{algorithmic}
\end{algorithm}
Combining the above equations in \textbf{lemmas 1-4}, we have the following relaxed problem without monotonicity constraint.
\begin{subequations}\label{opt3:max_Us}
    \begin{flalign}
    \max_{(R_n,e_n)}U_s&=\sum_{n=1}^Np_n(\lambda_1(\beta_1+\beta_2\theta_n-
    \beta_3e^{(-\beta_4{(10^{-3}e_n)}^{\beta_5})})\notag \\
    &+\lambda_2\ln{(T_{\max}-{T_n^{com}-\frac{e_nc_n}{fn}})}-\theta_nR_n)\label{opt3:objective}\\
    {\text{s.t.}}\ \  &\theta_1R_1-\xi{e}_1c_1f_1^2-E_1^{com} = 0\label{opt3:c1} \\
    &{\theta_n}R_n-\xi{e}_nc_nf_n^2 = {\theta_nR_{n-1}-\xi{e}_{n-1}c_{n-1}f_{n-1}^2},\notag \\
    &n=2, \cdots, N\\
    &T_{n}^{com}+\frac{e_{n}c_{n}}{f_{n}}\leq T_{\max}, n=1, \cdots, N.\label{opt3:c3}
    \end{flalign}
\end{subequations}

A standard method to resolve the original problem is to resolve the relaxed problem and the solution is then verified to satisfy the monotonicity condition. By iterating the constraints in the relaxed problem, we can get $R_n$ as
\begin{equation}\label{R_n}
\begin{aligned}
R_n=\sum_{i=2}^n\frac{1}{\theta_i}\xi c_if_i^2\left(e_i-e_{i-1}\right)+\frac{1}{\theta_1}\left(\xi c_1f_1^2e_1+E_1^{com}\right)
\end{aligned}
\end{equation}

By substituting $R_n$ into${\sum_{n=1}^N\theta_np_nR_n}$, we have
\begin{equation}\label{R_n_sum}
\begin{aligned}
\sum_{n=1}^{N}\theta_np_nR_n=\sum_{n=1}^{N}l_ne_n+\frac{E_1^{com}}{\theta_1}\sum_{n=1}^{N}\theta_np_n,
\end{aligned}
\end{equation}
where
\begin{equation}\label{l_n}
l_n=
\begin{cases}
{\xi c_nf_n^2p_n+\xi c_nf_n^2(\frac1{\theta_n}-\frac1{\theta_{n+1}})\sum_{i=n+1}^N\theta_ip_i}, & {n<N} \\
{\xi c_Nf_N^2p_N}, & {n=N}.
\end{cases}
\end{equation}

The result is brought into the objective function and converted into the optimization problem of a single variable $e_n$. We generate the function's graphical representation, the optimal efforts $e_n^*$ and the corresponding $R_n^*$ can be obtained. Considering the simplest market environment, the client levels follow uniform distribution, and the monotonicity is automatically satisfied \cite{13}. After the clients sign the contracts, the local model training epochs can be adaptively determined according to
\begin{equation}\label{tau_n}
{\tau_k=\left\lfloor\frac{e_{n}}{d_{k}}\right\rfloor },
\end{equation}
where the level of client $k$ is $n$, with 
$\theta_{k}\in(\theta_{n-1},\theta_{n}]$.
\begin{algorithm}[t]
\caption{Access Control Algorithm}
\label{alg:Access Control Algorithm}
\begin{algorithmic}[1] 
\FOR{$n=1, \cdots, N$}
    \FOR{$k\in level-n$}
        \IF{$|\bar{\mu}_{n}^t-\widehat{\mu}_{n}^t|>a$}
            \IF{$q_{k}^{t}<\bar{\mu}_{n}^t-\sigma_{n}^t$}
                \STATE $\mathcal{K}(t)\leftarrow\mathcal{K}(t)-\{k\}$;
            \ENDIF
        \ELSIF{$q_{k}^{t}<\bar{\mu}_{n}^t-\phi\sigma_{n}^t$}
            \STATE $\mathcal{K}(t)\leftarrow\mathcal{K}(t)-\{k\}$;
        \ENDIF
    \ENDFOR 
\ENDFOR
\STATE Set $Q\leftarrow\sum_{k\in\mathcal{K}(t)}q_{k}^{t}$;
\STATE Set $\alpha_{k}^{t}\leftarrow\frac{q_{k}^{t}}{Q}$;
\RETURN $\alpha_{k}^{t}$;
\end{algorithmic}
\end{algorithm}

\subsection{Access Control Algorithm}
At the end of a communication round, the clients ready to upload are classified according to their levels. For the clients of the same level, if one of them fails the access control algorithm, it cannot be added to the aggregation, and will not be paid the final reward. In the algorithm, we use learning quality, client level, and iteration delay time as access indicators, which can be used as aggregate weights. For the quantification of learning quality, one of the most direct indicators is the test accuracy of the local model. However, this will bring huge additional overhead for testing all local models' accuracy in the PS. Therefore, we used the loss reduction which can be calculated using the stored loss of local and global models, we define it as follows
\begin{equation}\label{m_k}
m_{k}^{t}=loss_{g}^{r}-loss_{k}^{t},
\end{equation}
where $loss_{g}^{r}$ denotes the test loss value of the global model for the time stamp index $r$ of client $k$ and $loss_{k}^{t}$ denotes the average training loss value of the client $k$ at round $t$. Meanwhile, combined with the levels and staleness, the access indicator of the client $k$ in the $t$-round aggregation is defined as
\begin{equation}\label{q_k}
q_{k}^{t}=
\begin{cases}
{m_{k}^{t}\cdot\theta_k\cdot(t-r+1)^{-\epsilon}}, &{k\in\mathcal{K}(t),} \\
{0}, &{k\notin\mathcal{K}(t)}.
\end{cases}
\end{equation}
where $\epsilon>0$ is the impact factor of staleness. We calculate the mean $\bar{\mu}_{n}^t$, median $\widehat{\mu}_{n}^t$, and standard deviation $\sigma_{n}^t$ of $q_{k}^{t}$ for the same clients level. Based on Chebyshev’s theorem,the access indicator tolerance $a$ is used, if $|\bar{\mu}_{n}^t-\widehat{\mu}_{n}^t|>a$ , the client with an access indicator less than $\bar{\mu}_{n}^t-\sigma_{n}^t$ will be prevented from entering the aggregation. Otherwise, the client with an access indicator less than $\bar{\mu}_{n}^t-\phi\sigma_{n}^t$ will be prevented from entering. After that, the weights are reset to complete the model aggregation. We finally get the model aggregation algorithm of \textbf{Algorithm \ref{alg:fedinasy}} and the access control algorithm of \textbf{Algorithm \ref{alg:Access Control Algorithm}}.

\begin{figure*}[!t]\centering
	\includegraphics[width=\linewidth]{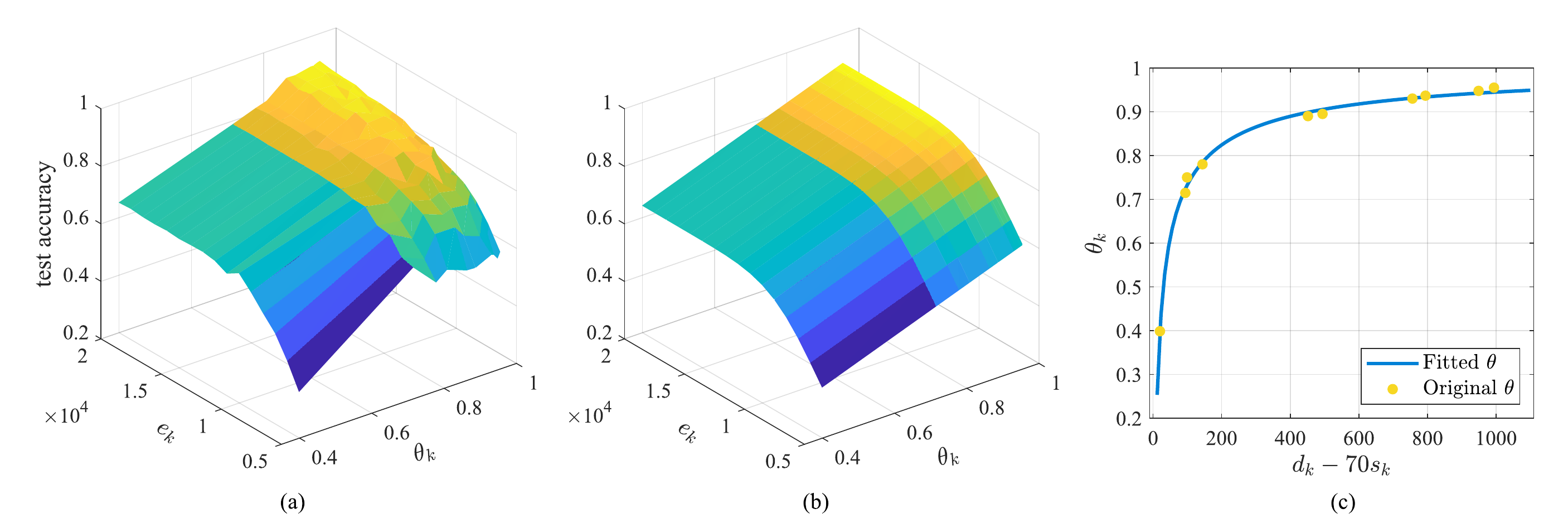}
	\caption{Experimental and fitting results: (a) experimental results of test accuracy; (b) fitting results of test accuracy; (c) fitting results of $\theta$.}
    \label{fig:nihe}
\end{figure*}

\section{Simulation Results}
\subsection{Verification utility function}
We use the MNIST dataset \cite{10} to carry out the test and determine the detailed parameters in \eqref{Eq:qet} by the curve fitting method. Without knowing the actual class distribution of the dataset, we assume that each label essentially has an equal probability of occurrence. Therefore, we set the benchmark for measuring the EMD value, to $\mathbb{P}(y~=~j)~=~0.1, j=0,\cdots,9$. Fig.~\ref{fig:nihe}ab shows the fitting results of the accuracy function on the MNIST dataset. The final fitting values are $\beta_1=0.459, \beta_2=0.432, \beta_3=0.459, \beta_4=0.009, \beta_5=2.436$.

\begin{figure}[!t]\centering
	\includegraphics[width=\linewidth]{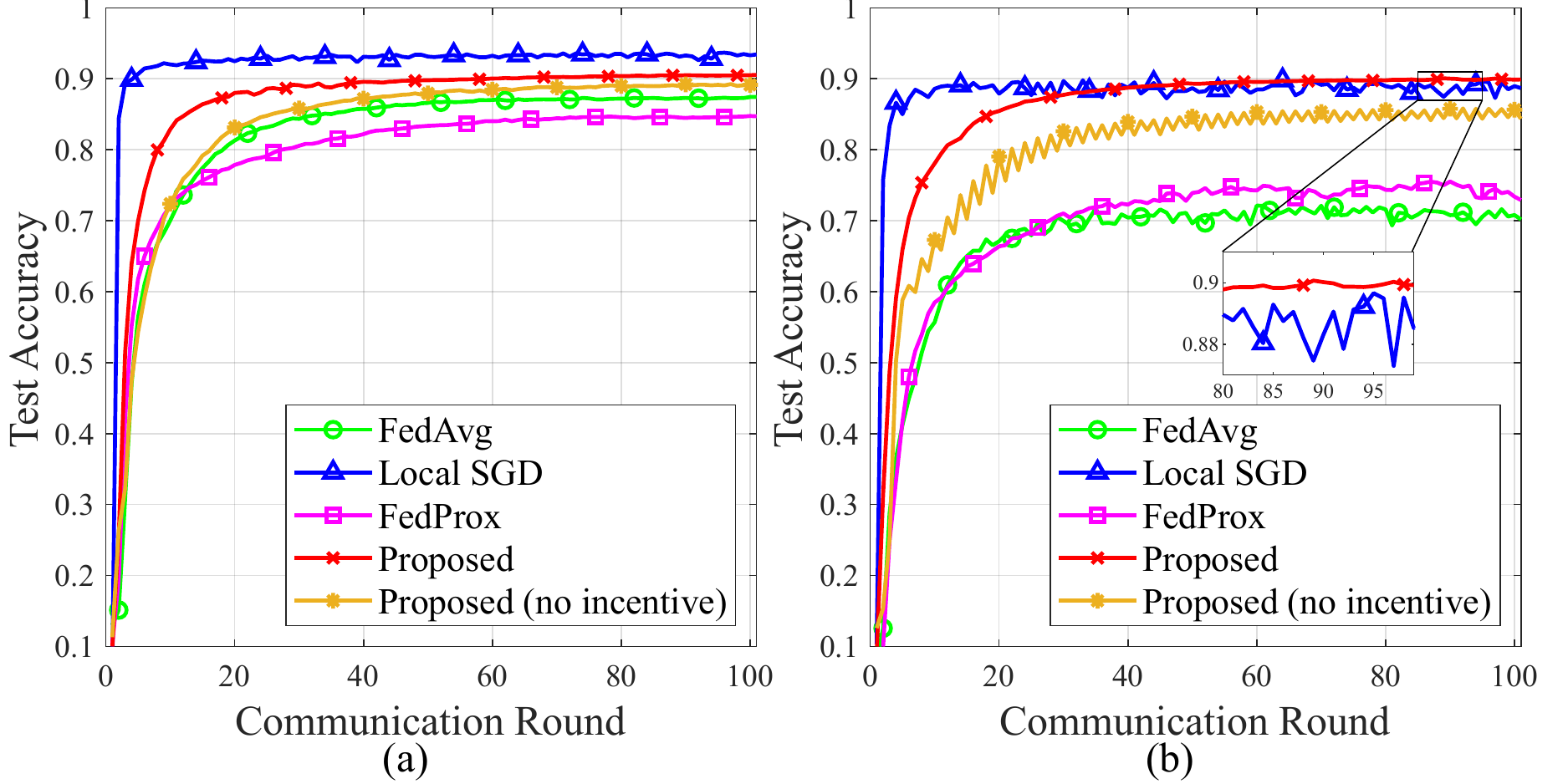}
	\caption{Test accuracy on experiment settings: (a) without attacks; (b) with 30  attackers.}
    \label{fig:test acc}
\end{figure}

The expression of $\theta$ in \eqref{Eq:theta} is determined according to the $\theta$ value obtained by the above fitting. In order to simplify the fitting function in the case of limited data points, we reduce the two-dimensional variables $d_k$ and $s_k$ to a single variable $d_k-70s_k$, aligning them on a comparable order of magnitude. Fig.~\ref{fig:nihe}c shows the experimental and fitting results of $\theta$ on the Mnist dataset. The final fitting values are $\gamma_1=10.559, \gamma_2=1.803, \gamma_3=70, \gamma_4=0.155$.

\subsection{Experiment Settings}
In the initial setup of a FL system, we assume 100 clients, 1 PS, and 1 task publisher. We trained a multi-layer perception (MLP) network with two hidden layers. In order to implement data heterogeneity, the data quantity for clients is set to follow the Zipf-1.0 distribution, the classes are set to follow the Dirichlet-0.1 distribution, and the final samples for each client belong to no more than 4 classes. We consider a set of data quality denoted by $\Theta=\{\theta_{1},\theta_{2},\ldots,\theta_{10}\}$ with 10 levels. the values follow uniform distribution $\mathcal{U}(0.1,1.0)$, where the clients with $\theta_k \in(\frac{n-1}{10},\frac n{10}]$ belong to level-$n$.

In the incentive mechanism, simulation parameters are set as follows: $\lambda_1=5000000, \lambda_2=400000, \xi=2, c_n=5, f_n=1, E_n^{com}=20, T_n^{com}=10, T_{max}=100000, \forall n\in\{1,...,10\}$. In the practical application of the asynchronous FL aggregation algorithm, due to the computation delay of each client following a uniform distribution $\mathcal{U}(0.5,2.0)$s, the period of model aggregation at each round is set to $\Delta t=1s$. Other experiment parameters are set as follows: batch-size $ =20$, learning rate $\eta=0.01$, $a=0.5$, $\epsilon=2.0$, and $\phi=3$.
\subsection{Performance Comparison}
We compare the performance of our proposed framework with the following FL algorithms:\textbf{(a) Local SGD}: An ideal synchronous FL algorithm where all the data in the market is concentrated in one client. \textbf{(b) proposed (no incentive)}: Asynchronous FL algorithm without incentive mechanism. \textbf{(c) FedAvg}\cite{6}: A synchronous aggregation baseline.\textbf{(d) FedProx}\cite{11}: Synchronous aggregation algorithm for dynamically adjusting the local training epochs.

In order to verify the effectiveness of the proposed framework, we assume that all clients are reliable, that is, there are no false sample labels. For synchronous FL, we set the local model training epochs to 10. Fig.~\ref{fig:test acc}a compares the test accuracy of all algorithms with respect to communication rounds. Observations reveal that our framework reached a test accuracy of 90.55\%, presenting a 2.88\% reduction compared to that of Local SGD. Our framework without the incentive mechanism achieved a test accuracy of 89.3\%, underscoring the improvement after integration. Furthermore, it demonstrated superior performance with a test accuracy 3.12\% higher than FedAvg and 5.84\% higher than FedProx. To verify the robustness against unreliable clients, we introduce 30 malicious clients to carry out attacks evenly distributed across 10 levels, and some of their training labels were deliberately modified. Fig.~\ref{fig:test acc}b compares the test accuracy of all algorithms with respect to communication rounds under attacks. It is evident that our framework achieved a notable test accuracy of 90.03\%, maintaining the performance in the presence of attacks, showcasing a 1.35\% increase compared to Local SGD. Our framework without incentive mechanism attained a respectable accuracy of 84.36\%. In contrast, the accuracy of FedAvg and FedProx significantly lagged behind, reaching only 70.18\% and 72.92\%, respectively. This can be attributed to the involvement of clients with malicious attacks, severely compromising the quality of the global model. Consequently, the task publisher withheld rewards of the 30 malicious clients.
\begin{table}[htbp]
\caption{Computation Time(s)}
\begin{center}
\begin{tabular}{c|c|c|c|c|c}
\hline
\multicolumn{2}{c|}{\textbf{Test Accuracy}}& \textbf{50\%} & \textbf{60\%} & \textbf{70\%} &\textbf{80\%}\\
\hline
\multirow{2}{*}{\textbf{Our proposed}} & without attacks & 2 & 3 & 4 & 8\\
\cline{2-6}
 & with 30 attackers  & 3 & 4 & 5 & 11\\
\hline
\multirow{2}{*}{\textbf{FedAvg}} & without attacks & 12.32 & 15.16 & 26.81 & 53.37\\
\cline{2-6}
 & with 30 attackers  & 23.04 & 36.32 & 92.74 & -\\
\hline
\multirow{2}{*}{\textbf{FedProx}} & without attacks & 16.97 & 22.65 & 45.36 & 147.40 \\
\cline{2-6}
 & with 30 attackers & 34.05 & 62.30 & 152.07 & -\\
\hline
\end{tabular}
\label{tab1}
\end{center}
\end{table}

From a computational time perspective, we present a list of all algorithms in Table \ref{tab1}. In experiments where computing capabilities are consistent, certain clients possessing significantly larger datasets compared to others. Consequently, they expend a considerable training time, exacerbating straggler issues during synchronous aggregation. However, our framework, integrating asynchronous aggregation and adaptive local training epochs, effectively mitigates straggler issues. As a result, the computation time is markedly reduced compared to other algorithms.

\section{Conclusion}
In this paper, we first introduce an innovative asynchronous federated learning framework, integrating contract-based incentive mechanisms to mitigate participation challenges and heterogeneity. By categorizing clients based on data quality and formulating contracts accordingly, we optimize the task publisher’s utility by adaptively adjusting the client’s local model training epochs. The proposed access control algorithm further enhanced security and accuracy. The simulation results substantiated the scheme's efficacy, showcasing its potential to stimulate high-quality model updates while preserving global accuracy and training efficiency. This approach holds promise for advancing federated learning in overcoming practical obstacles.


\begin{thebibliography}{00}
\bibitem{4}
Y.~Jiao, P.~Wang, D.~Niyato, B.~Lin, and D.~I. Kim, ``Toward an automated
  auction framework for wireless federated learning services market,''
  \emph{IEEE Trans. Mob. Comput.}, vol.~20, no.~10, pp. 3034--3048, 2021.

\bibitem{5}
J.~Kang, Z.~Xiong, D.~Niyato, S.~Xie, and J.~Zhang, ``Incentive mechanism for
  reliable federated learning: A joint optimization approach to combining
  reputation and contract theory,'' \emph{IEEE Internet Things J.}, vol.~6,
  no.~6, pp. 700--714, 2019.

\bibitem{14}
C.~Xie, O.~Koyejo, and I.~Gupta, ``Asynchronous federated optimization,''
  \emph{ArXiv}, vol. abs/1903.03934, 2019.

\bibitem{2}
C.-H. Hu, Z.~Chen, and E.~G. Larsson, ``Scheduling and aggregation design for
  asynchronous federated learning over wireless networks,'' \emph{IEEE J. Sel.
  Areas Commun.}, vol.~41, no.~4, pp. 874--886, 2023.

\bibitem{6}
B.~McMahan, E.~Moore, D.~Ramage, S.~Hampson, and B.~A.~y. Arcas,
  ``{Communication-Efficient Learning of Deep Networks from Decentralized
  Data},'' in \emph{Proceedings of the 20th International Conference on
  Artificial Intelligence and Statistics}, pp. 1273--1282, 2017.

\bibitem{8}
Y.~Liu, M.~Tian, Y.~Chen, Z.~Xiong, C.~Leung, and C.~Miao, \emph{A Contract
  Theory Based Incentive Mechanism for Federated Learning}, pp. 117--137, 2022.

\bibitem{13}
R.~Strausz, ``Bolton, p., and dewatripont, m.: Contract theory,'' \emph{Journal
  of Economics}, vol.~86, pp. 305--308, 2005.

\bibitem{10}
L.~Deng, ``The mnist database of handwritten digit images for machine learning
  research,'' \emph{IEEE Signal Process Mag.}, vol.~29, no.~6, pp. 141--142,
  2012.

\bibitem{11}
T.~Li, A.~K. Sahu, M.~Zaheer, M.~Sanjabi, A.~Talwalkar, and V.~Smith,
  ``Federated {{Optimization}} in {{Heterogeneous Networks}},'' in
  \emph{Proceedings of {{Machine Learning}} and {{Systems}}}, vol.~2, pp. 429--450, 2020.
\end{thebibliography}
\end{document}